\pdfoutput=1
\documentclass[a4paper]{cas-sc}

\usepackage[numbers]{natbib}
\usepackage{subcaption}
\usepackage{xcolor}

\def\tsc#1{\csdef{#1}{\textsc{\lowercase{#1}}\xspace}}
\tsc{WGM}
\tsc{QE}
\begin{document}
\let\WriteBookmarks\relax
\def\floatpagepagefraction{1}
\def\textpagefraction{.001}

\shorttitle{}    

\shortauthors{}  

\title [mode = title]{End-to-End Cell Detection via Instance-aware Graph Modeling}


\author[1]{Ruochen Liu}[orcid=0009-0008-5112-7489]

\ead{sgrliu18@liverpool.ac.uk}


\affiliation[1]{organization={Faculty of Science and Engineering, University of Liverpool},
            city={Liverpool},
%          citysep={}, % Uncomment if no comma needed between city and postcode
            postcode={L69 3GJ}, 
            country={U.K.}}

\author[2]{Yalin Zheng}[orcid=0000-0002-7873-0922]

% Email id of the second author
\ead{yzheng@liverpool.ac.uk}


\affiliation[2]{organization={Department of Eye and Vision Sciences, University
of Liverpool}, 
            city={Liverpool},
%          citysep={}, % Uncomment if no comma needed between city and postcode
            postcode={L7 8TX}, 
            country={U.K.}}

\author[3]{Jingxin Liu}[orcid=0000-0001-6071-9197]

% Email id of the second author
\ead{Jingxin.Liu@xjtlu.edu.cn}


\affiliation[3]{organization={School of AI and Advanced Computing, Xi’an Jiaotong-Liverpool University}, 
            city={Suzhou},
%          citysep={}, % Uncomment if no comma needed between city and postcode
            postcode={215400}, 
            country={China}}

\author[4]{Jianfeng Zhang}%[]

% Email id of the second author
\ead{jfzhang@zjnu.edu.cn}


\affiliation[4]{organization={College of Mathematical Medicine, Zhejiang Normal University}, 
            city={Jinhua},
%          citysep={}, % Uncomment if no comma needed between city and postcode
            postcode={321004}, 
            country={China}}

\author[4]{Shoujun Huang}%[]

% Email id of the second author
\ead{sjhuang@zjnu.edu.cn}


\author[5]{Dexing Kong}%[]

% Email id of the second author
\ead{dkong@zju.edu.cn}


\affiliation[5]{organization={School of Mathematical Sciences, Zhejiang University}, 
            city={Hangzhou},
%          citysep={}, % Uncomment if no comma needed between city and postcode
            postcode={310027}, 
            country={China}}

\author[6]{Haofeng Li}[orcid=0000-0001-9120-9843]
\cormark[1]

% Email id of the second author
\ead{lihf95@mail.sysu.edu.cn}


\affiliation[6]{organization={School of Systems Science and Engineering, Sun Yat-sen University}, 
            city={Guangzhou},
%          citysep={}, % Uncomment if no comma needed between city and postcode
            postcode={510275}, 
            country={China}}

\author[4]{Wei Lou}[orcid=0000-0002-2071-4081]

\cormark[1]

% Email id of the second author
\ead{louwei@zjnu.edu.cn}

            
\cortext[1]{Corresponding author}


\begin{abstract}
Accurate cell detection and classification are crucial for pathological analysis, directly affecting diagnostic accuracy and treatment planning. To capture complex cellular interactions beyond visual appearance within the tumor microenvironment, several approaches have employed graph neural networks to model spatial and relational patterns among cell nuclei, yielding promising results. However, these methods typically adopt a two-stage paradigm of visual extraction followed by relational modeling, which necessitates separate tuning for each stage, thereby increasing pipeline complexity and hindering end-to-end joint optimization. In this paper, we propose an end-to-end framework for cell detection and classification that jointly models patch-level visual representations and instance-level interactions, which incorporates a dynamic graph construction module and an instance-aware graph network. Specifically, the graph construction module dynamically builds the graph structure using learnable queries derived from patch-level features as cell instance representations, with adjacency defined by integrating feature similarity and spatial distances. The instance-aware graph network performs adaptive instance filtering and feature reorganization, aggregating them over the cell graph into a topological latent state for a selective state-space transition driven by visual cues, fusing appearance and relational evidence. When evaluated on multiple datasets with different staining protocols for cell and nucleus detection, our method significantly outperforms existing approaches in both detection and classification performance. The code will be released at https://github.com/RuochenLiu23/IGM.
\end{abstract}


%\item A unified end-to-end cell detection framework modeling inter-instance interactions.
%\item Learnable queries enable joint optimization of visual and relational modeling.
%\item A topology-structured SSM with visual features driving unified state evolution.

% Keywords

\maketitle

% Main text
\section{Introduction}\label{sec:intro}
Histopathological images are the gold standard for diagnosing and grading complex diseases such as cancer, playing a critical role in clinical diagnostics and public health~\cite{Baxi22}. Cell detection and classification, which involve identifying cell or nucleus locations and predicting their types, provide essential quantitative and qualitative support for diagnosis and prognosis~\cite{Pan18}. However, manual analysis is labor-intensive and costly due to the thousands of cells typically present in whole slide images (WSIs)~\cite{Sarah21,Cui21}. Although computer-aided methods enable rapid and accurate cell identification~\cite{Cui21}, their performance is challenged by dense clustering, extensive overlap, and adhesion that make adjacent instances difficult to separate~\cite{dist18,hover19,stardist18,donet}, ambiguous boundaries caused by uneven staining and imaging noise~\cite{dist18,huang23,micronet,donet}, and high cellular heterogeneity that manifests as intra-sample variability in morphology and staining intensity as well as differences across laboratory protocols~\cite{hover19,cellpose,jun24,pixel22,cellmamba}.

\begin{figure}
    \centering
    \includegraphics[width=0.5\textwidth]{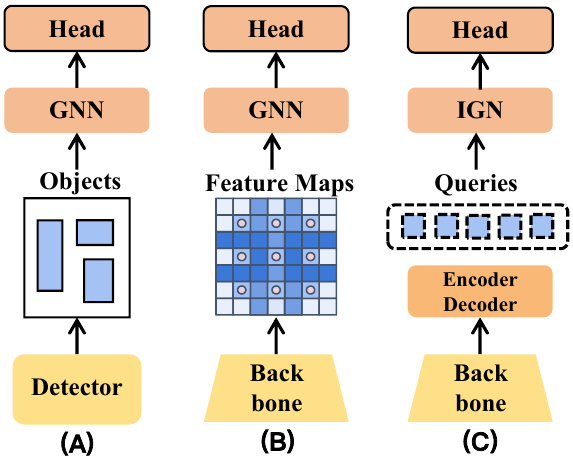}
    \caption{Comparison of different cell detection and classification approaches incorporating graph models: (A) Graph construction is initialized using the output of a pre-trained detector. (B) Graph construction is initialized based on the feature map. (C) Ours: Dynamic graphs are initialized using learnable queries. The Instance-aware Graph Network (IGN) performs adaptive noise filtering and feature reorganization, enabling effective capture of local‑global relationships.}
    \label{fig:framework_1}
\end{figure}

Extensive research has been dedicated to this challenging task, with cell and nucleus detection approaches broadly categorized into three groups: \textit{point-based}, \textit{instance segmentation}, and \textit{instance detection methods}~\cite{cellmamba}. \textit{Point-based methods}~\cite{sha21, shui22, dpap2p, towards25} predict only centroids and their categories to reduce computational complexity and annotation costs. However, they lack explicit regional information, making it challenging to recover cell shape or boundary details. This limits their ability to support downstream morphological analyses such as chromatin pattern characterization~\cite{stripenn,cancercells} or mitotic figure quantification~\cite{MIDOG2023,MIDOG2022}.  In contrast, \textit{instance segmentation methods}~\cite{dist18,hover19,stardist18,donet,cellpose,jun24,pixel22,cpp2023,cellvit} offer pixel-level delineation of individual cells but demand extensive annotations and high computational resources~\cite{nima2020}. \textit{Instance detection}~\cite{MIDOG2022,digestpath,zhu2024,cellmamba} strikes a favorable balance: it provides explicit regional localization via bounding boxes, with lower annotation and computational overhead than segmentation, while offering richer spatial detail than point-based approaches, making it well-suited for clinical deployment. Accordingly, we aim to develop an accurate and efficient framework for cell detection and classification.

Within the instance detection paradigm, query-based Transformer models~\cite{detr,dino,ddq,cellotype} excel by using learnable queries to represent objects. However, in cell detection, their reliance solely on visual and positional cues leads to ambiguity in dense, overlapping, or morphologically heterogeneous regions with high inter-class similarity. Pathology-specific methods~\cite{dist18,hover19,stardist18,cpp2023,cellvit} mitigate this by incorporating geometric priors (distance maps or contour fitting) but require pixel-level annotations and are incompatible with detection-only settings. With the rise of pathology foundation models, recent methods counter such ambiguity by enriching instance representation with pre-trained features, yet the objective gap between representation learning and coordinate regression degrades these features under joint optimization, necessitating decoupled multi-stage training~\cite{DeNuc}. In general, these approaches follow the conventional computer vision paradigm, modeling cells largely in isolation, prioritizing local appearance over inter-instance contextual relationships. This limits detection performance in complex tissue architectures where cellular identity is inherently relational. 

To address these challenges in pathology, previous studies have focused on intercellular interactions to provide additional cues, leading to the exploration of graph-based methods to model such contextual relationships in the tissue microenvironment. \textit{Two-stage approaches}~\cite{GNN2021,wei2024,taimur2022,GT2024,GrEp2026,relation2025} (Figure~\ref{fig:framework_1}(A)) first detect or segment cells and then build graphs from cropped regions, but suffer from error propagation from imperfect proposals and increased training complexity due to separate per-stage optimization. In contrast, \textit{pixel- or superpixel-based methods}~\cite{meng2021,gao2022} (Figure~\ref{fig:framework_1}(B)) construct graphs directly from feature maps with pixels as nodes, yet struggle to model meaningful inter-instance relationships while producing overly dense and noisy structures that obscure instance-level information.

Inspired by these observations, we propose an end-to-end framework to enhance instance awareness in query-based cell detection and classification. As shown in Figure~\ref{fig:framework_1}(C), our method treats learnable queries as graph nodes which represent candidate cell instances and construct a context-aware graph. The graph connectivity is determined by both spatial proximity and feature similarity, enabling dynamic adaptation to diverse cell distributions and morphologies. However, since the number of queries typically far exceeds the true cell count, the graph contains redundant and spurious nodes, introducing significant structural noise. This poses a major challenge for conventional GNNs~\cite{graph2017}, which rely solely on local message passing and thus struggle to capture long-range dependencies in such large, noisy graphs. Although Transformer-based GNNs~\cite{gat2018,TokenGT2022} can model global relationships, their dense attention mechanisms incur substantial computational overhead, especially when applied to noisy graphs with excessive nodes. Irrelevant nodes in such cases not only inflate computation cost but also distort attention, undermining inter-instance relationship modeling. To address this, we leverage the selective State Space Model (SSM)~\cite{mamba2024}, which effectively suppresses noise while preserving global structural patterns, thereby enabling robust relational modeling at linear computational cost. We embed the graph topology into the SSM, jointly modeling inter-cellular relations and visual representations.

In this paper, our framework comprises three key components: a query feature learning network, a dynamic graph construction module (DGC), and an instance-aware graph network (IGN). The query feature learning network fuses initial query features with tissue contextual and pathological features extracted from patches, producing enriched query embeddings that encode instance-level semantics, bridging patch-level features to instance-level representations. The DGC module takes the queries as graph nodes and dynamically constructs the graph structure based on both spatial distance and feature similarity among queries. This adaptive construction strategy allows the graph to better capture varying cell distributions and morphologies, enhancing the model's robustness during training. The IGN integrates two synergistic components: an Instance-aware Graph Learning (IGL) module, in which Selective Feature Reorganization (SFR) gates out redundant queries and reorganizes the retained ones into a directionally aligned latent space, whereupon graph aggregation superposes neighbors into near-orthogonal subspaces, yielding a noise-resistant topological representation; and a Topology-structured State Space (TSS) layer, where the state is transitioned along the cell-graph topology and query's visual feature drives state update, so that relational and visual evidence accumulate within a unified state. To account for the high semantic complexity of histopathological images, both IGL and TSS are augmented with a Context-Guided State Encoding module, which injects context-aware priors into dynamic state transitions. Finally, the refined node features from the TSS are decoded into cell instance predictions, including bounding box regression and class labels.

Our main contributions are summarized as follows:
\begin{enumerate}
    \item We propose a new end-to-end cell detection framework that leverages learnable queries for graph construction and relational modeling, aiming to enhance instance awareness and detection performance.
    \item We propose an instance-aware graph network that recasts graph learning as a topology-structured state space model, in which selective reorganization and orthogonal aggregation yield a noise-resistant topological structural representation as the latent state driven by visual evidence, fusing relational and appearance cues within the state transition.
    \item We propose a dynamic graph construction method that integrates both the spatial relationships and feature similarities of queries to build graphs and in turn adaptively adjusts their structure in response to dynamic updates in query content. 
    \item Experimental results demonstrate that our framework achieves state-of-the-art performance on multiple public benchmarks for cell and nucleus detection across different staining protocols.
\end{enumerate}

\section{Related Work}\label{sec:related}
\subsection{Instance Detection for Cells and Nuclei}
Cell detection and classification typically rely on instance segmentation or point detection. \textit{Segmentation methods}~\cite{stardist18,hover19,cellpose,cpp2023,cellvit,cisca2025} like Hover-Net~\cite{hover19} and StarDist~\cite{stardist18} provide detailed masks but require costly pixel-wise labels and are computationally expensive. \textit{Point-detection approaches}~\cite{sha21,shui22,dpap2p,towards25}, while efficient, only predict centroids and lack geometric information needed for morphometric analysis. \textit{Bounding-box-based instance detection}~\cite{MIDOG2022,digestpath,zhu2024,cellmamba} offers a balanced alternative, providing richer spatial localization than point detection while avoiding the overhead of full segmentation. Recent query-based detectors~\cite{detr,dino,ddq,cellotype,DeformableDetr,DabDetr,CoDetr,deco,attention2020,celldetr} such as Cell-DETR~\cite{celldetr} and DINO~\cite{dino,cellotype} enable end-to-end, anchor-free instance detection with strong global context modeling. However, these models have seen limited exploration in the field of pathology, as they primarily involve the adaptation of general-purpose models, leaving a need for more pathology-specific approaches. These limitations underscore the necessity of a cell instance detection framework.

\subsection{Graph-Based Learning in Pathology}
In computational pathology, graph-based cell modeling typically involves two stages: \textbf{graph construction} and \textbf{graph representation learning}. For \textbf{graph construction}, two strategies dominate: (1) \textit{Two-stage methods}~\cite{GNN2021,wei2024,taimur2022,GT2024,GrEp2026} first detect or segment cells to form semantically meaningful nodes and connect them via spatial proximity, but they suffer from error propagation and pipeline complexity. (2)\textit{ End-to-end pixel- or patch-level approaches}~\cite{gao2022} build graphs directly from feature maps using $k$-nearest neighbors (KNN), but often produce redundant or biologically irrelevant edges. For \textbf{graph representation learning}, Graph Convolutional Networks (GCNs)~\cite{graph2017,gasteiger2019,chen2020,eli2021} and Graph Attention Networks (GATs)~\cite{gat2018} are widely adopted. GCNs perform fixed-weight neighborhood aggregation but suffer from limited receptive fields and over-smoothing in deep architectures. GATs improve adaptivity by assigning attention-based weights to neighbors, but standard attention mechanisms incur quadratic computational complexity with respect to the number of nodes~\cite{attention2017}, which hinders scalability in dense cellular graphs containing thousands of instances. Recently, Mamba has been introduced into graph learning~\cite{KGMamba}, offering global modeling at linear complexity. Motivated by these observations, in this work, we propose an efficient dynamic graph learning approach that reformulates the state transition of a selective SSM over the cell graph topology, thereby achieving global contextual modeling of relational and visual representations at low complexity.

\section{Method}\label{sec:method}
\begin{figure}
    \centering
    \includegraphics[width=\textwidth]{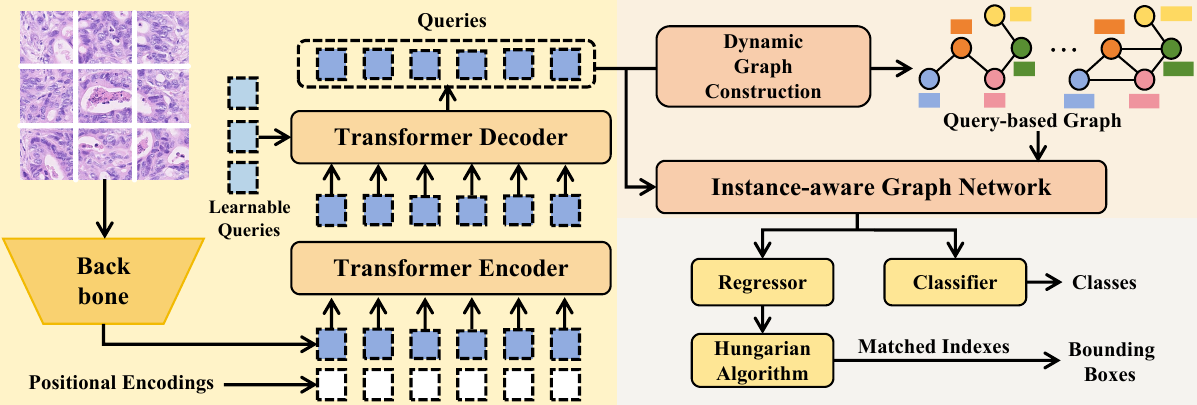}
    \caption{Overview of the whole framework. The proposed end-to-end learning framework comprises three key components: a query feature learning network for initializing query embeddings with patch-to-instance-level mapping, a dynamic graph construction (DGC) module for building the cell graph, and an instance-aware graph network (IGN) for effective graph modeling of inter-cell interactions, all jointly optimized during training.}
    \label{fig:framework2}
\end{figure}

In this section, we introduce the proposed end-to-end cell detection and classification framework, which is shown in Figure~\ref{fig:framework2}. The framework consists of three key components: (1) a query feature learning network that learns tissue and pathological features, extracts high-quality query representations from patches, achieving the mapping from patch-level to instance-level features; (2) a dynamic graph construction strategy that adaptively builds the cell graph by jointly leveraging spatial proximity and feature similarity among learnable queries; and (3) an instance-aware graph network that learns cell-level features by modeling local and global relational contexts with the visual evidence to enhance the discriminative ability of query features for both instance separation and class prediction.

\subsection{Query Feature Learning Network}
Given a pathological image of size $H \times W \times 3$, it is fed into a backbone network to obtain multi-scale visual features $f_{\text{multi}} = \{f_0, f_1, \dots, f_l\}$. These features are flattened and concatenated into an initial global feature $f \in \mathbb{R}^{N_{\text{enc}} \times C}$, where $N_{\text{enc}}$ is the total number of spatial positions across all scales. A 2D sinusoidal positional encoding is generated for each spatial location, flattened, and added to $f$, resulting in $f_{\text{enc}} \in \mathbb{R}^{N_{\text{enc}} \times C}$. The features are then processed by a Transformer encoder using multi-scale deformable attention~\cite{DeformableDetr}, which captures fine-grained pathological details and aggregates contextual information from the tissue microenvironment to produce enhanced visual representations $f'_{\text{enc}} \in \mathbb{R}^{N_{\text{enc}} \times C}$. From $f'_{\text{enc}}$, we generate initial object proposals. Specifically, each spatial position corresponds to a candidate box $\boldsymbol{a} = [c_x, c_y, w, h]$, where $(c_x, c_y)$ denote the center coordinates and $(w, h)$ are randomly initialized dimensions. After applying class-agnostic non-maximum suppression (NMS), the top-$N$ high-scoring proposals are selected. Their features are extracted from $f'_{\text{enc}}$ at $(c_x, c_y)$ and linearly transformed to form the anchor features $a_s \in \mathbb{R}^{N \times C}$.

Similar to common detection pipelines~\cite{DeformableDetr,dino,ddq}, for each anchor, a lightweight prediction head takes its associated feature to compute a score and a bounding-box regression offset, facilitating the selection of high-quality detection proposals: we first apply class-agnostic non-maximum suppression (NMS) on their boxes to remove duplicates, then keep the top-$N$ anchors by score ($N \gg$ the actual object count). To improve the quality of query embeddings, the selected anchor features $a_s \in \mathbb{R}^{N \times C}$ are added to a set of learnable query vectors $q' \in \mathbb{R}^{N \times C}$, yielding the final query features $Q_{\text{enc}} \in \mathbb{R}^{N \times C}$. These learnable query vectors serve as instance-level priors for potential objects (e.g., cells), which are randomly initialized and optimized during training but remain fixed during inference. The query features $Q_{\text{enc}}$, together with 4D reference points $(c_x, c_y, w, h)$ normalized to $[0,1]$, are fed into the Transformer decoder.

In the Transformer decoder, each decoder layer iteratively updates the query features. First, positional embeddings are generated from the current reference points using sine-based encoding followed by a MLP layer. Each query is updated through self-attention and multi-scale deformable cross-attention, enabling it to gather visual evidence from relevant regions across encoder feature scales. Owing to the dense sampling of initial queries, the receptive fields of queries overlap with each other. To further reduce redundancy and ensure the distinctness of queries for accelerating model convergence, between decoder layers, the model selects a subset of high-quality, non-overlapping queries using class-agnostic NMS with an IoU threshold of 0.8, based on the current 4D reference points and their classification scores.

Following the design of~\cite{ddq}, a subset of queries determined by the configuration is treated as dense and routed to the auxiliary head for refinement to ensure the integrity of feature representations, while they are not used for final predictions at inference. In addition, 100 denoising (DN) queries are added only during training as noisy copies of ground-truth boxes and labels to provide direct supervision and speed up convergence. Finally, the Transformer decoder outputs the updated query features $Q$ and predicts the offset of the selected anchor positions. Combining these offsets with the original anchor allows us to calculate the updated reference points, as well as the extraction of the center coordinates $P \in \mathbb{R}^{N \times 2}$, providing key spatial cues for subsequent cell graph construction and relationship modeling.

\subsection{Dynamic Graph Construction}
Existing graph-based cell classification methods typically use K-Nearest Neighbor (KNN) based on spatial proximity to construct cell graphs~\cite{GNN2021,wei2024,taimur2022,gao2022}. However, in our end-to-end framework, the 2D coordinates of reference points are not always reliable. Furthermore, solely spatial information has limited capacity to model complex intercellular distributions and interactions, rendering such constructions sensitive to noise and harmful to convergence. To address this, we propose a Dynamic Graph Construction (DGC) strategy that combines both query embeddings and positional information to define a more robust query distance function in KNN, enabling accurate relationship modeling. Notably, by constructing the graph over queries, whose features and reference points are dynamically updated during training, the cell graph structure is refined via gradient propagation through query features. This dynamic refinement enhances flexibility and avoids the limitations of fixed graph topologies.

Based on the query embeddings $Q$ and reference points $P$ output by the query feature learning network, we design a query distance function $s$ for KNN computation, which consists of two components: spatial distance $s^p$ and feature distance $s^f$. For spatial distance, the distance is defined via an exponential decay of the Euclidean distance between any two reference points $p_i$ and $p_j$:
\begin{equation}
    s_{ij}^p = \exp\left( -\frac{\| p_i - p_j \|_2}{2\bar{d}} \right)
\end{equation}
where $\| p_i - p_j \|_2$ is the Euclidean distance between $p_i$ and $p_j$, and $\bar{d}$ denotes the mean distance between all pairs of reference points. A large $s^p$ indicates stronger spatial correlation.

For feature distance, we first perform L2 normalization on the query embeddings $Q$ to eliminate the influence of feature scale. Subsequently, we measure feature distance $s_{ij}^f$ using cosine similarity between any two queries $q_i$ and $q_j$.  The formula is as follows:
\begin{equation}
s_{ij}^f = \frac{q_i \cdot q_j^T}{\lVert q_i \rVert_2 \, \lVert q_j \rVert_2}
\end{equation}
In the end, the query distance function can be described as:
\begin{equation}
s_{ij} = \alpha \cdot (1-s_{ij}^p) + (1-\alpha) \cdot (1-s_{ij}^f)
\end{equation}
where $\alpha$ is a hyper-parameter that controls the contribution weights of the two distances. For dynamic balance in this specific setting, $\alpha$ is set to 0.5. Then, we apply the KNN algorithm to establish node connectivity based on the query distance $s_{ij}$, resulting in an adjacency matrix $M \in \mathbb{R}^{N \times N}$, where $n$ is the number of queries. The constructed graph is defined as $G = (V, E)$, where each node $v_i \in V$ corresponds to a query embedding $q$, and edges $E$ are determined by the KNN connections.

\subsection{Instance-aware Graph Network}
\begin{figure} 
    \centering
    \includegraphics[width=\textwidth]{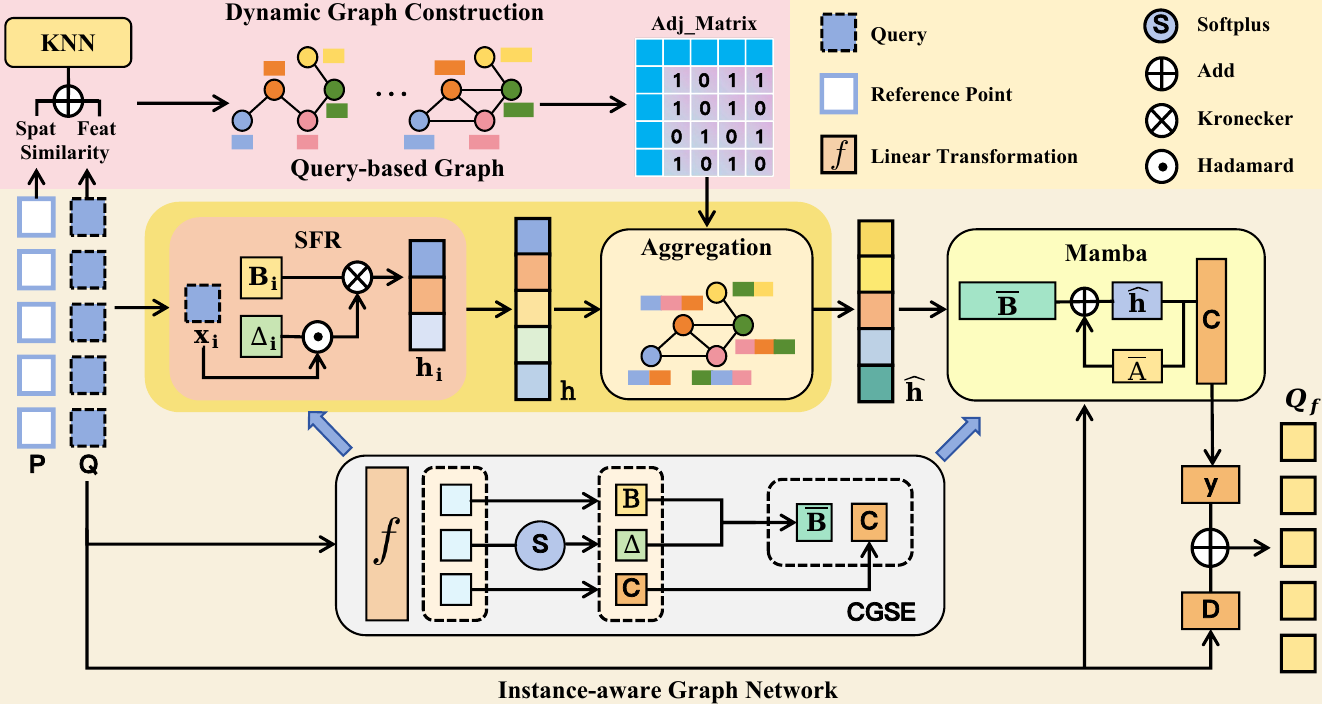} 
    \caption{The proposed Dynamic Graph Construction (DGC) strategy and Instance-aware Graph Network (IGN) are employed to construct and model the query-based graph for simulating inter-cell interactions.}
    \label{fig:framework3}
\end{figure}

We propose an Instance-aware Graph Network (IGN) for dynamic query-based cell detection, as illustrated in Figure~\ref{fig:framework3}. IGN comprises two tightly coupled components: the Instance-aware Graph Learning (IGL) and the Topology-structured State Space (TSS) layer. In IGL, Selective Feature Reorganization (SFR) applies a selective mechanism to suppress redundant query features and reorganizes the retained features into a directionally structured latent space. The resulting representations are then aggregated over the dynamic cell graph, where near-orthogonal directions suppress interference from semantically inconsistent neighbors, yielding a noise-resistant topological representation of semantic associations. Within TSS, the state is updated over this representation under the drive of instance-level query features, integrating contextual information with global relational dependencies. A Context-Guided State Encoding (CGSE) module abstracts all queries into an instance-level context that steers the state-transition parameters of both SFR and TSS, guiding adaptation to instance-specific feature distributions.

\subsubsection{Instance-aware Graph Learning}
Within this module, selective feature reorganization is coupled with graph aggregation. SFR first gates out redundant queries and lifts the retained ones into a directionally structured latent space, rendering dissimilar cells near-orthogonal. Aggregating these over the dynamic cell graph then consolidates the orthogonalized node representations with graph connectivity into a unified topological structural representation, wherein inconsistent neighbors reside in mutually orthogonal subspaces and remain non-interfering. Similarity-driven selective aggregation therefore arises from the state geometry itself, obviating explicit edge pruning and rendering the structure robust to redundant nodes.

\noindent{\textbf{Selective Feature Reorganization.}}
Given a set of nodes $V$, we flatten their node features into a sequence $x\in \mathbb{R}^{N \times d}$ and feed it into the Selective Feature Reorganization (SFR) module. Inspired by the selective state space model~\cite{mamba2024}, we apply a selective projection to adaptively filter and reconstruct each node feature $x_i\in \mathbb{R}^{d}$, where the projection matrices $\Delta \in \mathbb{R}^{N \times d}$ and $\mathbf{B} \in \mathbb{R}^{N \times n}$ are dynamically determined by Context-Guided State Encoding, yielding the hidden state $h_i\in \mathbb{R}^{d \times n}$. 
\begin{equation}
h_i = \left(\boldsymbol{\Delta} \odot x_i \right) \otimes \mathbf{B}_i
\end{equation}
where $\otimes$ denotes the outer product.

$\Delta_i \in \mathbb{R}^{d}$ serves as an element-wise amplitude gate that controls the amount of information written from each node feature into the state space. By implicitly encoding the node information through CGSE, $\Delta$ achieves content-adaptive modulation, enabling effective filtering and compression of noisy node features in a fully data-driven manner.

$\mathbf{B}_i \in \mathbb{R}^{n}$ expands each scalar component of $x_i \in \mathbb{R}^{d}$ onto the direction it spans, yielding $h_i \in \mathbb{R}^{d \times n}$. Lifting node features transforms the entangled space into a directionally discriminative one, mapping similar node vectors $h_i$ to proximal directions and dissimilar ones to near-orthogonal directions, whereby the affinity between two states factorizes as $\cos(h_i,h_j)=\cos(\tilde{x}_i,\tilde{x}_j)\cdot\cos(\mathbf{B}_i,\mathbf{B}_j)$ with $\tilde{x}_i=\boldsymbol{\Delta}_i \odot x_i$, sharpening the contrast between related and unrelated node pairs.

\noindent{\textbf{Directional Graph Aggregation.}}
Building on these near-orthogonal features, we aggregate them over the cell graph, injecting graph connectivity to form a topological representation. We remap all $h_i\in \mathbb{R}^{d \times n}$ as node features $h \in \mathbb{R}^{N \times (d \cdot n)}$ into a cell graph $G$, whose connectivity is established using the adjacency matrix $M\in \mathbb{R}^{N \times N}$ derived from the dynamic graph construction strategy. To balance node degree differences and enhance stability, we first perform symmetric normalization on it~\cite{graph2017}, followed by aggregating adjacent features based on structural information. The relevant formulas are as follows:
\begin{equation}
\hat{h} = \mathcal{D}^{-1/2} \cdot (M + I) \cdot \mathcal{D}^{-1/2} \cdot h
\end{equation}
where $I$ denotes the identity matrix for adding self-loops, $\mathcal{D}$ represents the degree matrix, and the two $\mathcal{D}^{-1/2}$ terms perform symmetric normalization on the target and neighbor nodes respectively. Node-wise, each node takes a degree-normalized weighted sum of its own state and those of its neighbors, whereby the graph connectivity determines which neighbors
are included and the normalization determines how much each contributes.

Since each node's state is confined to the direction assigned by its own $\mathbf{B}_i$, this weighted sum acts as a superposition along distinct directions rather than a blending of values. Semantically inconsistent neighbors, whose states point in near-orthogonal directions, therefore enter the aggregated state without contaminating the target node's own state and remain separable at readout. Selectivity thus resides in the geometry rather than in the edges, requiring no explicit pruning, and the aggregated $\hat{h}$ constitutes a noise-resistant topological structural representation consolidating node features with graph connectivity.

\subsubsection{Topology-structured State Space}
Due to the complexity and global nature of graph structures, we employ Mamba~\cite{mamba2024} to model global features. Specifically, we propose the Topology-structured State Space (TSS) layer, which adopts the topological structure as the latent state and drives its update with the visual cues of the queries, thereby constructing a topology-structured state transition that yields the output $y$:
\begin{equation}
H = \bar{\mathbf{A}} \hat{h} + \bar{\mathbf{B}} x, \quad y = \mathbf{C} H
\end{equation}
In contrast to a canonical state space update, driven by the state of the preceding token, here the cell topology $\hat{h}$ obtained by graph aggregation serves as the incoming state, with the visual features $x$ continually re-injected to drive each update. Instance-level appearance is thereby assimilated into the topological structure, unifying relational and appearance evidence within a single state $H$.

Since Mamba is defined on the continuous time, Zero-Order Hold (ZOH) discretization is applied to adapt the continuous-time model to discrete node sequence inputs and facilitate training:
\begin{equation}
\bar{\mathbf{A}} = \exp(\Delta \mathbf{A}), \quad
\bar{\mathbf{B}} = (\Delta \mathbf{A})^{-1} \bigl( \exp(\Delta \mathbf{A}) - \mathbf{I} \bigr) \cdot \Delta \mathbf{B}
\end{equation}
Herein, $\mathbf{A}$ and $\mathbf{B}$ are all parameter matrices, which are discretized based on the parameter $\Delta$ and their specific parameters are determined by CGSE.

Unlike Transformer, which explicitly computes pairwise similarities at quadratic complexity to obtain the global state, Mamba implicitly encodes global instance and topological context into a compact hidden state $H$. Transformer attention weights, by contrast, tend to disperse over complex graph structures, leaving irrelevant node signals difficult to suppress. Within TSS, $\bar{\mathbf{A}}$ adaptively determines, conditioned on the representation of the corresponding instance, how much of the accumulated topological structure survives each update, so that the contextual topology aggregated onto visually substantiated cells persists. $\bar{\mathbf{B}}$ writes the visual feature of each query into the very state that carries the topology, along the direction spanned by $\mathbf{B}_i$ and scaled to the salience of that instance, thereby driving the state forward at every step. $\mathbf{C}$ performs the inverse operation, projecting the fused state back onto content-dependent directions to retrieve for each node its relevant global context, which keeps the readout selective at linear complexity. Meanwhile, cells admit no canonical ordering within tissue and thus form an unordered set rather than a sequence. We process nodes in parallel and keep the topological propagation permutation-equivariant, thereby avoiding the ordering bias inherent to sequential formulations.

To enhance feature propagation and enrich representational capacity, a learnable feedforward matrix $D$ is introduced to form a residual-style connection, yielding the IGN output $Q_f = y + \mathbf{D} x_0$.

The fused features $Q_f$ are fed into the classification head to predict the object categories. Simultaneously, $Q_f$ and its corresponding reference point are used by the regression head to predict the bounding box coordinates. Finally, predictions are matched to ground truths via the Hungarian algorithm, and the overall loss is computed accordingly.
\begin{equation}
\mathcal{L} = \lambda_{\text{cls}} \mathcal{L}_{\text{cls}} + \lambda_{\text{bbox}} \mathcal{L}_{\text{bbox}} + \lambda_{\text{iou}} \mathcal{L}_{\text{iou}}
\end{equation}
where $\mathcal{L}_{{bbox}}$ is the L1 Loss, $\mathcal{L}_{{iou}}$ is the GIoU Loss~\cite{GIoU}, and $\mathcal{L}_{{cls}}$ employs Focal Loss~\cite{FocalLoss} to address the positive-negative sample imbalance in pathological images.

\subsubsection{Context-Guided State Encoding.}
In SSM, $\mathbf{A}$, $\mathbf{B}$, $\mathbf{C}$, and $\boldsymbol{\Delta}$ are the key matrices governing the state transition. We propose a Context-Guided State Encoding (CGSE) module to generate these matrices from instance features, enabling adaptive, knowledge-guided node control.

The state transition matrix $\mathbf{A}$ is initialized via HiPPO-LegS~\cite{hippo2020}, providing a stable multi-scale decay spectrum for task-relevant retention and noise suppression during hidden state propagation. As $\mathbf{A}$ encodes task-level global decay structure, it is shared globally across all nodes as a single learnable parameter. For matrices $\mathbf{B}$ and $\mathbf{C}$~\cite{mamba2024, mbagcn2025}, we employ a linear transformation to learn knowledge $\mathbf{W}_x$ from the cell instance information. Following the tensor generation mechanism used in self-attention~\cite{attention2017}, this cellular knowledge, encompassing the morphological appearance and spatial distribution of cell instances, is transformed through learned
mappings into corresponding matrices, enabling adaptive processing of nodes guided by all cell instance features. The calculation process is as follows:
\begin{equation}
\mathbf{B} = f(x, \mathbf{W}_x) \mathbf{W}_B, \quad \mathbf{C} = f(x, \mathbf{W}_x) \mathbf{W}_C 
\end{equation}
For the $\Delta$, we use an activation function Softplus~\cite{mamba2024} to adapt the characteristics of the modulation matrix:
\begin{equation}
    \boldsymbol{\Delta} = \log\left(1 + \exp({f(x, \mathbf{W}_x) \cdot \mathbf{W}_{\delta}} \cdot \mathbf{W}_{dt})\right)
\end{equation}
Due to the large output dimensionality of $\Delta$, a low-rank structure $\mathbf{W}_{\delta} \cdot \mathbf{W}_{dt}$ with $\mathbf{W}_{\delta}\in\mathbb{R}^{d\times r}$ and $\mathbf{W}_{dt}\in\mathbb{R}^{r\times d}$ ($r\ll d$), is additionally employed to control the parameter count.

\section{Experiment}\label{sec:exp}
\subsection{Datasets}
We evaluated our method on three public histopathological datasets: CoNSeP~\cite{hover19}, CytoDArk0~\cite{cisca2025}, and OCELOT~\cite{ocelot}, with annotations converted to bounding box labels. CoNSeP comprises 41 H\&E-stained 1000×1000-pixel images of colorectal adenocarcinoma at 40× magnification with nucleus-level instance segmentation masks. CytoDArk0 includes Nissl-stained mammalian brain tissue images at 40× magnification with cell-level masks for neurons and glial cells. OCELOT is derived from H\&E-stained WSIs in the TCGA database, with cell nuclei annotated. To balance efficiency and resolution, 128×128 patches are extracted from CoNSeP due to its denser cell distribution, while 256×256 patches are used for CytoDArk0 and OCELOT. Evaluations on nuclear detection, cell detection, and multi-type datasets with diverse staining methods confirm the model’s adaptability across pathological detection tasks.

\begin{table*}[!htb]
\centering
\caption{Performance comparison of different methods on instance detection tasks across three datasets.}
\resizebox{0.7 \textwidth}{!}{
\begin{tabular}{c|c|c|c|c|c}
\toprule
\textbf{Dataset} & \textbf{Methods} & \textbf{AR} & \textbf{mAP} & \textbf{mAP$_{50}$} & \textbf{mAP$_{75}$} \\
\midrule
\multirow{10}{*}{\textbf{CoNSeP}} 
& Deformable-DETR (ICLR'21) & 49.3 & 24.5 & 50.3 & 25.2   \\
& DAB-DETR (ICLR'22) & 49.6 & 24.9 & 50.5 & 25.5   \\
& DINO (ICLR'23) & 54.8 & 28.5 & 52.4 & 28.9  \\
& CO-DINO (ICCV'23) & 53.3 & 29.8 & \underline{54.6} & 29.2   \\
& DDQ-DETR (CVPR'23) & \underline{56.5} & \underline{30.0} & 54.4 & \underline{31.4}   \\
& Relation-DETR (ECCV'24) & 55.1 & 29.7 & 54.3 & 30.3   \\
& Mamba-YOLO (AAAI'24) & 51.3 & 25.2 & 50.7 & 23.6  \\
& DECO (ICLR'25) & 53.5 & 26.9 & 51.4 & 25.7   \\
& CellMamba (BMVC'25) & 53.2 & 25.7 & 51.1 & 23.8   \\
& Ours & \textbf{58.7} & \textbf{32.0} & \textbf{57.0} & \textbf{33.5}   \\
\midrule
\multirow{10}{*}{\textbf{CytoDArk0}}
& Deformable-DETR (ICLR'21) & 67.5 & 52.1 & 80.1 & 59.3   \\
& DAB-DETR (ICLR'22) & 67.8 & 52.6 & 80.5 & 59.4   \\
& DINO (ICLR'23)& 69.9 & 58.2 & 84.7 & 67.1   \\
& CO-DINO (ICCV'23) & 70.0 & 59.0 & 85.6 & 68.7   \\
& DDQ-DETR (CVPR'23) & \underline{70.3} & \underline{59.6} & \underline{86.3} & \underline{69.5}   \\
& Relation-DETR (ECCV'24) & 70.1 & 58.8 & 85.9 & 68.2   \\
& Mamba-YOLO (AAAI'24) & 67.4 & 52.3 & 81.2 & 56.6   \\
& DECO (ICLR'25) & 68.9 & 54.6 & 81.9 & 63.1   \\
& CellMamba (BMVC'25) & 68.2 & 53.3 & 83.5 & 59.8   \\
& Ours & \textbf{71.7} & \textbf{61.7} & \textbf{88.5} & \textbf{71.8}   \\
\midrule
\multirow{10}{*}{\textbf{OCELOT}} 
& Deformable-DETR (ICLR'21) & 54.5 & 31.5 & 68.4 & 24.8   \\
& DAB-DETR (ICLR'22) & 54.7 & 31.2 & 68.9 & 24.4   \\
& DINO (ICLR'23) & 55.2 & 33.3 & 71.0 & 27.3   \\
& CO-DINO (ICCV'23) & 54.9 & \underline{33.5} & 71.0 & \underline{27.5} \\
& DDQ-DETR (CVPR'23) & \underline{55.5} & \underline{33.5} & \underline{71.1} & 27.2   \\
& Relation-DETR (ECCV'24) & 55.1 & 32.8 & 70.6 & 26.0  \\
& Mamba-YOLO (AAAI'24) & 54.7 & 30.9 & 68.2 & 23.6   \\
& DECO (ICLR'25) & 55.0 & 32.1 & 68.8 & 25.9 \\
& CellMamba (BMVC'25) & 54.3 & 31.7 & 69.0 & 25.4   \\
& Ours & \textbf{56.8} & \textbf{34.9} & \textbf{72.3} & \textbf{29.3}  \\
\bottomrule
\end{tabular}
}
\label{tab:different_datasets_methods_detection_performance}
\end{table*}

\subsection{Implementation Details}
All experiments were conducted on an NVIDIA A40 GPU. We adopt ResNet-50 as the backbone, ensuring that the performance advantage stems from the detection architecture design itself rather than from backbone capacity. For training, we employed the AdamW optimizer with an initial learning rate of 2e-4 and a weight decay coefficient of 0.05. For the hyperparameter \( K \) in KNN adopted in Dynamic Graph Construction, experimental validation demonstrates that setting \( K = 8 \) on CoNSeP, \( K = 3 \) on CytoDArk0, and \( K = 4 \) on OCELOT achieves the optimal performance. To ensure fairness in the comparison, we unified the number of queries to 900 for general DETR-based methods. 

\subsection{Evaluation Setup and Metrics}
To comprehensively evaluate our model, we compare against both detection and classification methods. The comparison with detection methods spans general-purpose and pathology-specific models: general-purpose detectors, though not designed for histopathology, have shown strong capability on pathological tasks, while pathology-specific models are largely adapted from such frameworks, and both are therefore included for a thorough validation. We then compare with classification methods on the multi-class dataset, allowing us to assess our classification performance against pathology-specific models. In terms of metrics, we adopt standard detection metrics including Average Precision ($mAP$), $mAP_{50}$, $mAP_{75}$, and Average Recall ($AR$) to assess detection and classification performance for detection methods across confidence thresholds. For comparison with cell-specific classification methods, which are mostly supervised with masks or points and differ in output granularity, we follow~\cite{celldetr} and adopt the F1-score, the standard metric for classification, as a consistent measure across methods.

\subsection{Comparison with the Existing Methods}
\subsubsection{Comparison with Detection Methods}
We evaluate our framework on CoNSeP, CytoDArk0, and OCELOT against state-of-the-art detectors. Specifically, Deformable-DETR~\cite{DeformableDetr} enhances feature learning; DAB-DETR~\cite{DabDetr}, DINO~\cite{dino}, and DDQ-DETR~\cite{ddq} improve query design; CO-DINO~\cite{CoDetr} introduces more comprehensive supervision; Relation-DETR~\cite{rdetr} exploits attention-based query-feature interaction; while Mamba-YOLO~\cite{Mambayolo}, DECO~\cite{deco}, and CellMamba~\cite{cellmamba} prioritize computational efficiency by leveraging the strengths of Mamba and convolution. As reported in Table~\ref{tab:different_datasets_methods_detection_performance}, our method achieves top performance across all three benchmarks. On CoNSeP, it surpasses DDQ-DETR~\cite{ddq} by 2.0\% in $mAP$ and 2.2\% in $AR$, the most notable improvement among the three benchmarks. CoNSeP is also the most challenging, with dense nuclei and high inter-class similarity hindering both instance separation and type discrimination, explicitly modeling inter-cell relations showing a clear advantage over appearance-driven detection here. On CytoDArk0, the margin over DDQ-DETR~\cite{ddq} is 2.1\% in $mAP$ and 1.4\% in $AR$. On OCELOT, consistent superiority is maintained across all metrics. All improvements are statistically significant ($p<0.05$). These results demonstrate that modeling inter-cell relationships via our instance-aware graph network provides critical relational cues absent in general-purpose detectors, validating its efficacy for pathological cell instance detection.

\begin{figure}
    \includegraphics[width=1\linewidth]{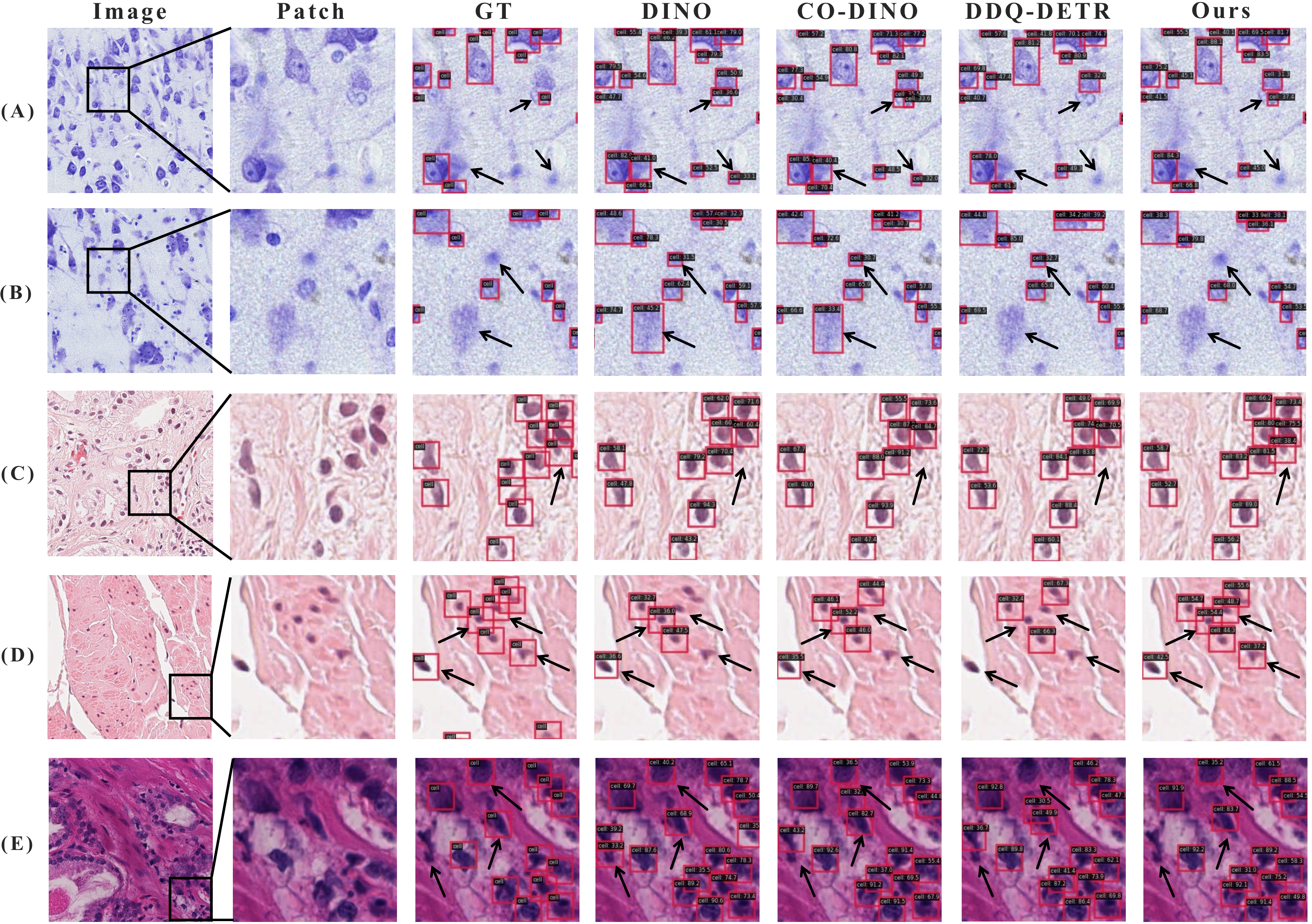}
    \caption{Visualization of cell detection results.}
    \label{fig:vis}
\end{figure}

We also visualize detection results from DINO, CO-DINO, DDQ-DETR, and our model in Figure~\ref{fig:vis} across different staining protocols, where (A) and (B) are from Nissl staining, and (C)--(E) are from H\&E staining. The results confirm that our method detects more cells while reducing false positives, particularly for densely packed and adherent cells.

\subsubsection{Comparison with Classification Methods}
Since most existing nucleus classification methods rely on segmentation masks or point annotations, we further compare our detection-based approach against representative end-to-end mask-based and point-based cell instance classification models. As shown in Table~\ref{tab:point_mask_cls}, we select a set of representative models for comparison, including DIST~\cite{dist18}, Micro-Net~\cite{micronet}, Hover-Net~\cite{hover19}, Triple-UNet~\cite{TripleU-net}, MCSpatNet~\cite{sha21}, TSFD-Net~\cite{TSFD}, Mask2Former~\cite{mask2former}, PGT~\cite{PGT}, ACFormer~\cite{huang23}, SMILE~\cite{smile}, CellViT~\cite{cellvit}, Cell-DETR~\cite{celldetr} and PathContext~\cite{towards25}. Our method achieves state-of-the-art performance across all metrics, outperforming mask-based methods with richer pixel-level supervision and point-based methods with inherently lower task complexity. These results validate the effectiveness of our detection-based framework for accurate cell localization and classification, particularly under severe class imbalance in pathological images. Notably, ${F_c}^m$ varies most substantially across methods, with several failing to identify this class entirely, as its scarcity and atypical morphology afford few reliable appearance cues. Our method attains the highest ${F_c}^m$, suggesting that contextual dependencies among neighboring cells yield discriminative evidence beyond individual appearance.

\begin{table}[htbp]
\centering
\caption{Performance comparison of nuclei classification methods and our approach on the CoNSeP dataset, reporting detection, mean and per-class F1 scores. $^{\dagger}$ Mask-based methods. $^{\ddagger}$ Point-based methods. $^{\star}$ Bounding Box-based methods.}
\resizebox{\textwidth}{!}{
\begin{tabular}{c|c|c|c|c|c|c|c|c|c|c|c|c|c}
\toprule
Model & $F_d$ & $F_{avg}$ & ${F_c}^m$ & ${F_c}^i$ & ${F_c}^e$ & ${F_c}^s$ & Model & $F_d$ & $F_{avg}$ & ${F_c}^m$ & ${F_c}^i$ & ${F_c}^e$ & ${F_c}^s$ \\
\midrule
DIST$^{\dagger}$ (TMI'18)& 0.71 & 0.42 & 0.00 & 0.53 & 0.62 & 0.51 & PGT$^{\ddagger}$ (MICCAI'23) & 0.74 & - & - & 0.62 & 0.64 & - \\
Micro-Net$^{\dagger}$ (MIA'19)& 0.74 & 0.47 & 0.12 & 0.59 & 0.62 & 0.53 & ACFormer$^{\ddagger}$ (ICCV'23) & 0.74 & - & - & 0.64 & 0.64 & - \\
Hover-Net$^{\dagger}$ (MIA'19)& 0.75 & 0.57 & 0.43 & 0.63 & 0.64 & 0.57 & SMILE$^{\dagger}$ (MIA'23) & \underline{0.76} & 0.56 & 0.38 & 0.62 & \textbf{0.67} & \underline{0.58} \\
Triple-UNet$^{\dagger}$ (MIA'20)& 0.66 & 0.38 & 0.10 & 0.57 & 0.42 & 0.44 & CellViT$^{\dagger}$ (MIA'24) & \underline{0.76} & 0.58 & 0.46 & 0.64 & \underline{0.65} & 0.57 \\
MCSpatNet$^{\ddagger}$ (ICCV'21)& 0.73 & 0.51 & 0.40 & 0.54 & 0.58 & 0.54 & Cell-DETR$^{\dagger}$ (MIDL'24) & 0.74 & 0.49 & 0.21 & 0.63 & 0.61 & 0.51 \\
TSFD-Net$^{\dagger}$ (NN'22) & 0.68 & 0.44 & 0.12 & 0.57 & 0.56 & 0.51 & PathContext$^{\ddagger}$ (AAAI'26) & \underline{0.76} & \underline{0.59} & \underline{0.49} & \underline{0.65} & \underline{0.65} & 0.57 \\
Mask2Former$^{\dagger}$ (CVPR'22) & 0.66 & 0.41 & 0.33 & 0.46 & 0.46 & 0.41 & Ours$^{\star}$ & \textbf{0.77} & \textbf{0.61} & \textbf{0.52} & \textbf{0.66} & \textbf{0.67} & \textbf{0.59}\\
\bottomrule
\end{tabular}
}
\label{tab:point_mask_cls}
\end{table}

\subsection{Ablation Study}
We conduct ablation studies on the CoNSeP dataset to validate the design choices of our framework, as reported in Table~\ref{tab:model_variant_performance}. We ablate two key components: the Dynamic Graph Construction and the Instance-aware Graph Network (IGN). First, we evaluate the importance of dynamic graph construction by ablating either feature similarity or positional similarity, using only one cue for graph building while keeping all other components fixed. Removing positional similarity reduces $mAP$ from 32.0\% to 31.4\% (-0.6\%) and $AR$ from 58.7\% to 57.3\% (-1.4\%). Removing feature similarity causes a larger drop: $mAP$ falls to 30.6\% (-1.4\%) and $AR$ to 56.6\% (-2.1\%), confirming that both cues are essential, where feature similarity contributes more substantially to overall performance. Second, we validate the effectiveness of each component within IGN. Starting from the baseline ($mAP$: 30.0\%, $AR$: 56.5\%), integrating the Instance-aware Graph Learning (IGL) module improves $mAP$ by 0.8\%, and further incorporating the Topology-structured State Space (TSS) yields an additional 1.2\% gain. To further verify the overall effectiveness of IGN, we replace it with alternative graph learners: the convolution-based GCN~\cite{graph2017}, the Transformer-based GAT~\cite{gat2018} and TokenGT~\cite{TokenGT2022}, and the Mamba-based DMbaGCN~\cite{DMbaGCN}. Our IGN outperforms all variants by clear margins, demonstrating its superiority.

\begin{table}
\centering
\caption{Ablation study on CoNSeP dataset.}
\begin{tabular}{l|c|c|c|c}
\toprule
Model Variant & AR & mAP & mAP$_{50}$ & mAP$_{75}$ \\
\midrule \midrule
\multicolumn{5}{l}{Ablation of Dynamic Graph Construction} \\
\midrule
w/o Feat. Sim. & 56.6 & 30.6 & 54.9 & 31.8 \\
w/o Pos. Corr. & 57.3 & 31.4 & 55.5 & 32.9 \\
Ours & \textbf{58.7} & \textbf{32.0} & \textbf{57.0} & \textbf{33.5} \\
\midrule \midrule
\multicolumn{5}{l}{Ablation of Instance-aware Graph Network} \\
\midrule
Baseline (Query Feature Learning Network) & 56.5 & 30.0 & 54.4 & 31.4 \\
+ $\text{IGL}$ & 57.4 & 30.8 & 56.1 & 32.3 \\
+ $\text{TSS}$ (Ours) & \textbf{58.7} & \textbf{32.0} & \textbf{57.0} & \textbf{33.5} \\
GCN (ICLR'17)& 56.3 & 30.2 & 54.6 & 31.7 \\
GAT (ICLR'19)& 56.7 & 30.5 & 55.0 & 32.0 \\
TokenGT (NeurIPS'22)& 56.9 & 31.1 & 55.7 & 32.6 \\
DMbaGCN (AAAI'26) & 57.2 & 30.9 & 55.5 & 32.1 \\
\bottomrule
\end{tabular}
\label{tab:model_variant_performance}
\end{table}

\begin{table}
\centering
\caption{Computational complexity comparison. $^{\dagger}$ Mask-based methods. $^{\ddagger}$ Point-based methods. $^{\star}$ Bounding Box-based methods.}
\begin{tabular}{c|c|c}
\toprule
Method & Para. (M) & FLOP (G) \\
\midrule
Hover-Net$^{\dagger}$~\cite{hover19} & 34.76 & 1992  \\
Mask2Former$^{\dagger}$~\cite{mask2former} & 298.21 & 896  \\
CellViT$^{\dagger}$~\cite{cellvit} & 142.85 & 3567  \\
DINO$^{\star}$~\cite{dino} & 47.55 & 279  \\
CO-DINO$^{\star}$~\cite{CoDetr} & 65.49 & -  \\
DDQ-DETR$^{\star}$~\cite{ddq} & 48.32 & 270 \\
PathContext$^{\ddagger}$~\cite{towards25} & 48.08 & 186 \\
Ours$^{\star}$ & 49.57 & 276  \\
\bottomrule
\end{tabular}
\label{tab:computational_complexity}
\end{table}

\subsection{Efficiency analysis}
We compare the computational complexity of our method with representative models on the CoNSeP dataset, as summarized in Table~\ref{tab:computational_complexity}. Our model (49.57M parameters, 276 GFLOPs) is substantially lighter than instance segmentation methods and comparable to point-based models, yet accomplishes a more complex task by providing instance-level representations approximating segmentation quality via bounding box detection. The marginal overhead over DDQ-DETR (+1.25M parameters, + 6 GFLOPs) confirms that our graph-based refinement module introduces minimal additional cost, achieving a favorable trade-off between efficiency and detection performance.

\subsection{Investigation of Hyper-parameters}
\subsubsection{Investigation of $\alpha$-values}
In constructing the query distance function, we incorporate both spatial distance and feature distance, balanced by a weighting factor $\alpha$. Therefore, we perform a hyperparameter study on $\alpha$ to determine its optimal setting. As shown in Figure~\ref{fig:alpha_ablation}, the optimal performance is achieved at $\alpha = 0.5$, indicating that spatial distance and feature distance contribute equally and complementarily to the query distance function. Accordingly, $\alpha = 0.5$ is adopted in our model.

\subsubsection{Investigation of Query Numbers}
\begin{figure}
    \centering
    \begin{subfigure}[b]{0.48\linewidth}
        \centering
        \includegraphics[width=\linewidth]{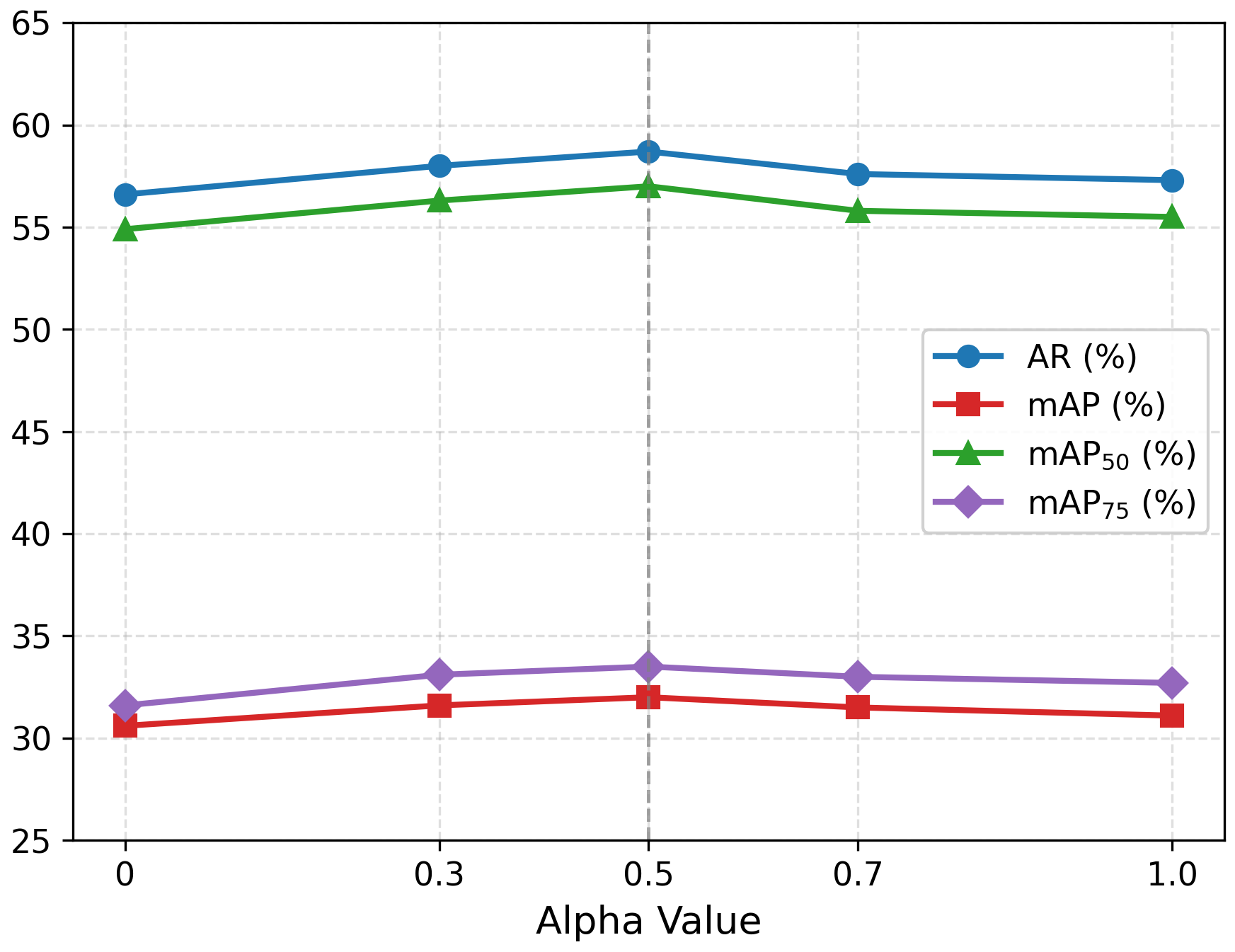}
        \caption{Different $\alpha$-values.}
        \label{fig:alpha_ablation}
    \end{subfigure}
    \hfill
    \begin{subfigure}[b]{0.48\linewidth}
        \centering
        \includegraphics[width=\linewidth]{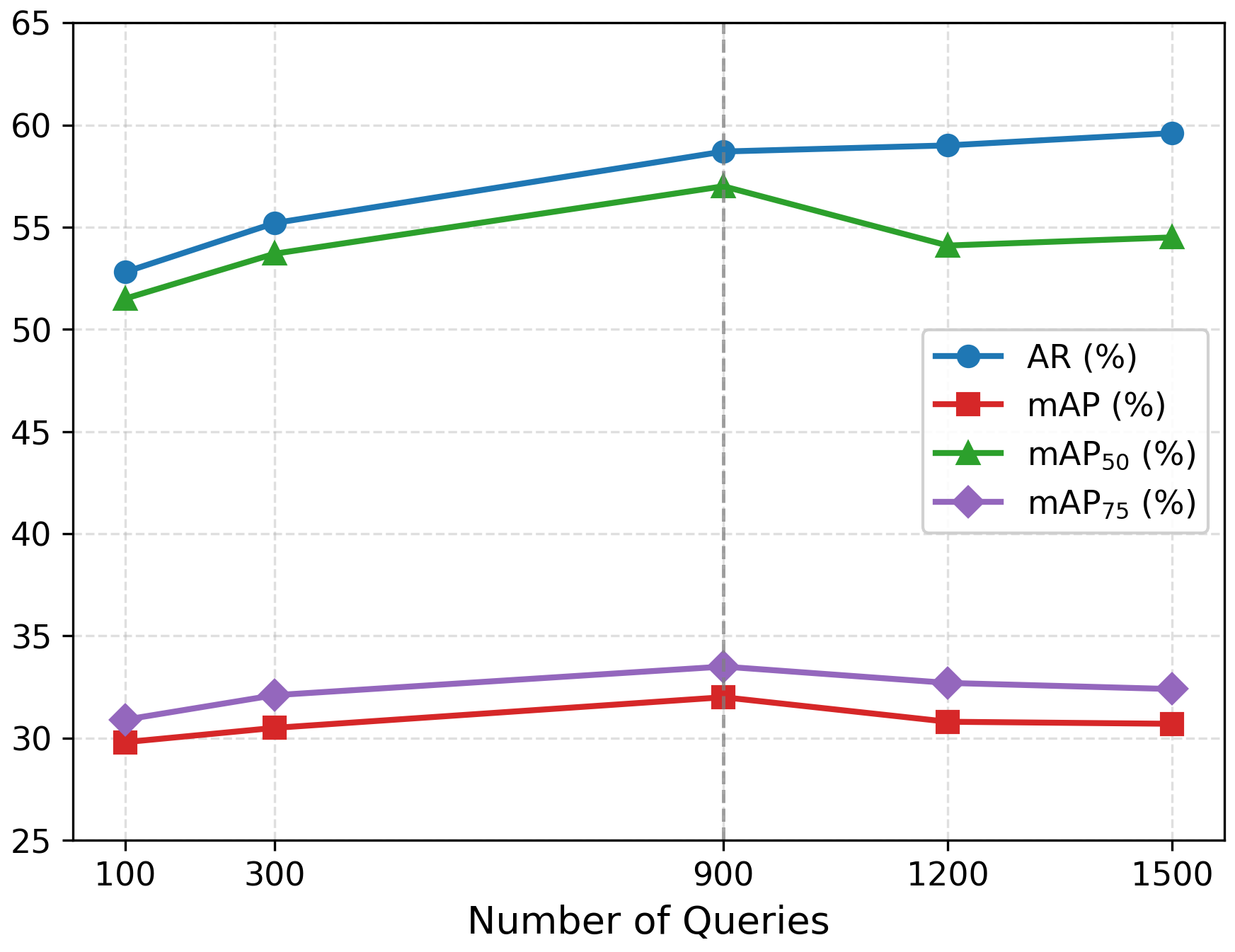}
        \caption{Different query numbers.}
        \label{fig:query_ablation}
    \end{subfigure}
    \caption{Hyper-parameter studies on CoNSeP.}
    \label{fig:hyperparam_ablation}
\end{figure}

The number of queries plays a critical role in query-based detectors, as it directly determines the upper bound on the number of detectable instances. Therefore, we conduct a dedicated hyperparameter study on query numbers. As shown in Figure~\ref{fig:query_ablation}, increasing $N$ consistently improves performance. Raising $N$ from 100 to 300 boosts $AR$ by 2.3\% and $mAP_{50}$ by 1.9\%; further increasing to 900 yields additional gains of 3.8\% and 2.3\%, respectively, indicating better candidate coverage and feature querying. When the number of queries is further increased, while AR improves, precision metrics decrease. We thus adopt 900 as the number of queries for our model to achieve the optimal balance between recall and precision. Larger values were not tested due to GPU memory constraints.

\subsubsection{Investigation of K-values}
In handling complex graphs, the graph construction step is of critical importance: the graph topology directly governs the neighborhood aggregation, which in turn conditions the subsequent global modeling. Herein, we utilize KNN to establish node connectivity, and the setting of K value is particularly vital to graph construction. Therefore, we conduct additional experiments on this hyperparameter K.

Based on empirical observations, the selection of $K$ can be broadly categorized into two regimes: sparse graph connectivity ($K \in [3, 5]$) and dense graph connectivity ($K \in [7, 9]$). We first conduct pilot experiments with $K=5$ and $K=8$ as representative values of each regime to determine which connectivity pattern is more suitable for each dataset. Based on these results, we further evaluate adjacent $K$ values within the identified regime to pinpoint the optimal setting.

As summarized in Table~\ref{tab:different_k_values_detection_performance}, on CoNSeP where nuclei are densely packed and highly overlapping, the best performance is achieved at $K=8$, outperforming sparser graphs (e.g., $K=5$) by up to 2.5\% in $mAP$. In contrast, CytoDArk0 contains well-separated cells, and its peak performance occurs at $K=3$, while OCELOT achieves the best results at $K=4$.

\begin{table*}
\caption{Performance comparison of different K values on detection tasks across three datasets.}
\centering
\begin{tabular}{c|c|cccc}
\toprule
\textbf{Dataset} & \textbf{K values} & \textbf{AR (\%)} & \textbf{mAP (\%)} & \textbf{mAP$_{50}$ (\%)} & \textbf{mAP$_{75}$ (\%)} \\
\midrule
\multirow{4}{*}{\textbf{CoNSeP}} 
& 5 & 54.7 & 29.5 & 54.5 & 30.4 \\
& 7 & 56.5 & 31.3 & 56.7 & 32.3 \\
& 8 & \textbf{58.7} & \textbf{32.0} & \textbf{57.0} & \textbf{33.5} \\
& 9 & 55.8 & 30.9 & 55.8 & 32.0 \\
\midrule
\multirow{4}{*}{\textbf{CytoDArk0}}
& 3 & \textbf{71.7} & \textbf{61.7} & \textbf{88.5} & \textbf{71.8} \\
& 4 & 71.3 & \textbf{61.1} & 88.1 & \textbf{70.8} \\
& 5 & 70.4 & 60.8 & 87.7 & 70.7 \\
& 8 & 68.5 & 57.6 & 86.7 & 67.9 \\
\midrule
\multirow{4}{*}{\textbf{OCELOT}} 
& 3 & 56.5 & 34.8 & 72.1 & 29.1 \\
& 4 & \textbf{56.8} & \textbf{34.9} & \textbf{72.3} & \textbf{29.3} \\
& 5 & 56.4 & 34.1 & 71.8 & 28.3 \\
& 8 & 56.6 & 33.5 & 70.9 & 27.1 \\
\bottomrule
\end{tabular}
\label{tab:different_k_values_detection_performance}
\end{table*}

\section{Discussion}\label{sec:discussion}
The central insight of this work is that cellular identity in histopathology is inherently relational. Intercellular interactions and signaling constitute the biophysical basis of structural integration and functional coordination in multicellular systems, such that the state of a cell is shaped not merely by its own morphology but largely by its interactions within the surrounding microenvironment. Existing detectors, however, model cells largely in isolation and prioritize local appearance, an approach that falters in dense, overlapping, or morphologically heterogeneous regions where visual cues alone cannot resolve instance identity. Certain pathology-specific methods instead introduce relational context through graph structures, yet predominantly follow a two-stage paradigm that decouples graph construction from detection: the graph is built upon the outputs of an independently trained detector or segmentor, serving only as post-hoc refinement, propagating upstream errors and precluding optimization under a unified objective. Neither line of work treats relational structure as something to be learned jointly with detection, which motivates a framework in which relational reasoning is not appended to detection but intrinsic to it.

To realize an end-to-end graph-based detection framework, we reconceive what constitutes a graph node. Rather than deriving nodes from the outputs of an upstream detector or from pixels on a feature map, we take learnable queries as candidate cell instances and construct the graph directly over them. Through decoding, each query aggregates multi-scale visual evidence from the tissue microenvironment, thereby transforming patch-level contextual representations into instance-level semantic descriptions upon which graph construction and relational reasoning operate. A single representation thus spans the entire path from patch-level feature extraction, through instantiation, to inter-instance relational modeling, allowing the graph to focus on instances themselves and all three stages to be optimized under a unified objective. Connectivity is determined jointly by spatial proximity and feature similarity, enabling the graph to encode organizational regularities that neither cue expresses alone. The resulting structure is genuinely dynamic: as query embeddings evolve through learning, the induced adjacency shifts accordingly; conversely, gradients propagated through graph learning continually refine the query features themselves, so that representation and topology shape each other over training and erroneous connections arising from unreliable reference points or immature features are not permanently entrenched. The topology is thereby shaped dynamically by the data rather than fixed a priori by spatial heuristics.

We next turn to how such a graph ought to be processed. As the number of queries far exceeds that of true cells, the graph abounds in redundant nodes, over which dense attention merely disperses irrelevant signals at quadratic cost. The state transition of a selective state space model, by contrast, is inherently endowed with content-adaptive selectivity, suppressing irrelevant signals while preserving global structural patterns at a complexity that scales linearly with the number of nodes. Departing from the denoise-then-model paradigm, we let graph learning itself yield a structure that the state space model can carry: once selectively reorganized, node states unfold along their respective directions, whereby aggregation becomes a superposition across distinct directions and semantically inconsistent neighbors, being near-orthogonal, remain mutually non-interfering. Conventional denoising schemes must assess neighbor relevance through attention weights or edge pruning, whereas filtering here follows directly from directional relations: irrelevant neighbors require no separate identification or removal, their contribution having attenuated substantially over end-to-end training. Building upon this, we recast the state transition of the SSM, taking the resulting topological representation as the latent state while the visual features of the queries are re-injected to drive its evolution, so that relational and appearance evidence accumulate within a single state. CGSE further abstracts all queries into an instance-level context from which the transition parameters are generated, enabling the dynamics to adapt to the feature distribution of each individual instance rather than conforming to a uniformly imposed prior.

Within the overall architecture, neither NMS nor KNN, despite being non-differentiable, impedes end-to-end training. Both act as parameter-free structural selectors conditioned on the current query states: NMS deterministically retains high-quality, non-redundant candidates, while KNN establishes connectivity from the prevailing query features. Neither introduces learnable parameters nor interrupts any gradient path, and the features of the retained queries participate fully in backpropagation. The dynamic character of the graph arises not from any adjustment within KNN itself, but from the continual learning of the query embeddings upon which it operates, thereby establishing an implicit feedback loop between feature learning and graph topology that constitutes the core mechanism of end-to-end joint optimization throughout the framework.

\section{Conclusion}\label{sec:conclusion}
In this paper, we have presented a novel end-to-end framework for cell detection and classification, designed to enhance the instance discrimination capability of query-based methods through structured reasoning. By treating learnable queries as graph nodes, our approach dynamically constructs a context-aware topology using both spatial and feature cues, enabling adaptive modeling of complex and varying cell distributions. To address noise and long-range dependency challenges in large query sets, we introduce an instance-aware graph network (IGN) that formulates a Topology-structured State Space (TSS) model with graph topology as the latent state, driving global modeling. Extensive experiments demonstrate the effectiveness of our method, achieving state-of-the-art performance across multiple benchmarks. This work highlights the potential of combining query-based detection with structured graph learning, offering a promising direction for precise and robust cell instance analysis in digital pathology.

\bibliographystyle{cas-model2-names}

\bibliography{cas-refs}



\end{document}